\documentclass[12pt]{article}

\usepackage{xcolor}
\usepackage{graphicx}

\begin{document}

\title{Parallel Corpus Augmentation using Masked Language Models}
\author{Vibhuti Kumari \and Kavi Narayana Murthy\\School of Computer and Information Sciences,\\University of Hyderabad\\vibhuroy9711@gmail.com,knmuh@yahoo.com}

\date{}

\maketitle

\bibliographystyle{plain}

\begin{abstract}

  In this paper we propose a  novel method of augmenting parallel text
  corpora which promises good quality and is also capable of producing
  many fold larger  corpora than the seed corpus we  start with. We do
  not need  any additional monolingual corpora.   We use Multi-Lingual
  Masked  Language Model  to  mask and  predict  alternative words  in
  context and we use Sentence  Embeddings to check and select sentence
  pairs which are  likely to be translations of each  other.  We cross
  check  our  method using  metrics  for  MT Quality  Estimation.   We
  believe this method can greatly  alleviate the data scarcity problem
  for  all  language pairs  for  which  a  reasonable seed  corpus  is
  available.
  
\end{abstract}

\section{Introduction}

Our focus  here is Machine  Translation, in particular  Neural Machine
Translation  (NMT). Deep  Learning Architectures  such as  Transformer
\cite{vaswani2017} have  given us NMT  systems that produce  very high
quality translations  compared to  any of the  other methods  known so
far,   provided   large   scale   parallel   corpora   are   available
\cite{Koehn}. A parallel  corpus includes two plain  text files, named
say F1  and F2, one  in language L1 and  another in language  L2, such
that  each line  in these  files contains  one segment  and the  lines
exhibit  parallelism  in  sense of  translational  equivalence.   Each
segment  is typically  one sentence,  but  it can  also be  part of  a
sentence or even  two or more sentences where needed  to depict proper
translational  equivalence.  First  line in  F1 and  first line  in F2
should be translations of each other, and so on for all the lines.\\

NMT systems produce  good translations only if  large parallel corpora
are available. NMT is data  hungry \cite{jain}. Large parallel corpora
are  not  always  available,  because  manual  translation  or  manual
checking   and  editing   of  MT   outputs   is  too   slow  and   too
costly. Parallel corpora  are usually built by  searching the Internet
for sentence pairs  that are likely to be translations  of each other,
based on certain statistical  measures of similarity. Parallel corpora
generated by such automated methods  do have quality issues. Even with
all this,  parallel corpora of  adequate quality in  adequate quantity
are  often   not  available,  leading   us  to  the  idea   of  corpus
augmentation.\\

Corpus  augmentation deals  with techniques  of starting  with a  seed
corpus and  automatically generating new sentences  from the sentences
in this seed  corpus by making small changes.  Corpus augmentation has
been widely studied  in the field of Image Processing  as also to some
extent in Natural  Language Processing (NLP).  Early  attempts at text
corpus augmentation such as  EDA \cite{eda} involve random insertions,
random deletions,  random swaps etc.   which can seriously  affect the
syntax and semantics of the  sentences. Such methods may be acceptable
for  certain  tasks such  as  Text  Classification  but they  are  not
suitable for tasks  such as Machine Translation where  it is important
to  preserve the  structure, meaning  and fluency.  Also, most  of the
works  done   in  text   corpus  augmentation  deal   with  augmenting
monolingual   corpora.   Here   our  focus   is  on   parallel  corpus
augmentation.  Very  little work  has  been  done in  parallel  corpus
augmentation. Preserving  parallelism makes this task  inherently more
difficult.\\

Most of the methods proposed for parallel corpus augmentation leverage
extra  monolingual  data   through  back-translation  \cite{Sennrich},
self-training  \cite{He}, or  translation  knowledge transfer  through
parallel    data   involving    other    assisting   language    pairs
\cite{firata,firatb,Nguyen}. In  this paper we propose  a novel method
for  parallel corpus  augmentation which  does not  require additional
monolingual corpora. We  use masked language models to  mask words and
predict alternatives which fit the context. This way we can generate a
large number of sentences from a given seed sentence and hopefully the
sentences  so generated  are  syntactically and  semantically of  good
quality. We take a pair of sentences,  let us call them s1 and s2, one
in L1 and the other in L2,  generate new sentences from each of these,
then  check for  translational  equivalence  using sentence  embedding
scores. If  highly reliable word alignment  techniques were available,
we  could restrict  this checking  process to  sentences generated  by
masking  aligned  words  only.   However, no  alignment  technique  is
dependable enough  today for the said  purpose and we need  to compare
all  sentences generated  from s1  with all  sentences generated  from
s2. Overall, we will be able to produce many new sentence pairs from a
given pair of sentences.\\

In   particular,    we   use   XLM-RoBERTa   model    \cite{xlmr},   a
Transformer-based BERT  model, in  the Fill-Mask mode.   The Fill-Mask
task involves masking and predicting which words to replace the masked
words.  Alexis et al.  \cite{xlmr}  claim that this method gave better
performance  as compared  to  other multilingual  language models  for
monolingual data. We then use  Sentence Transformers such as the LaBSE
Sentence  Embedding  model  to  check for  similarity.  We  check  the
sentences  generated in  two languages  pairwise for  for their  LaBSE
score  and filter  in  only those  pairs  which get  a  score above  a
specified threshold.

\section{Related Work}

Data  Augmentation involves  generating synthetic  data from  existing
data   in   order   to   increase  the   amount   of   training   data
\cite{Iterative}.  It  is an  effective  technique  for improving  the
performance  of Machine  Learning models.  Data augmentation  has been
studied more extensively  in computer vision than  in natural language
processing  \cite{Connor}.   In  image  datasets,   data  augmentation
involves  applying transformations  such  as  rotation, cropping,  and
changing  the  brightness   of  images.  For  text   data,  Easy  Data
Augmentation (EDA) \cite{eda} was one of the first works. EDA involves
operations  such  as  random  insertion, random  deletion  and  random
swapping of  words. This  can significantly  affect the  syntactic and
semantic  quality of  the  generated corpus.  Other text  augmentation
methods   include   back-translation  \cite{hayashi},   synonym   word
substitution \cite{Georgios}. Instead of replacing words with synonyms
from  a  specific  source,  using word  embeddings  such  as  Word2Vec
\cite{mikolov}  words   can  be  replaced  with   their  most  similar
counterparts  \cite{marivate}.   Transformer  based   text  generation
techniques such as GPT-3 \cite{Brown} and BERT \cite{kenton} have also
been proposed.\\

Jason  et al.   \cite{eda}  present  a simple  set  of universal  data
augmentation   techniques    for   NLP    called   EDA    (Easy   Data
Augmentation). EDA consists  of four simple yet  powerful operations -
synonym  replacement,  random  insertion,   random  swap,  and  random
deletion. In  synonym replacement  (SR) , we  randomly choose  n words
from the  sentence that are not  stop words. We replace  each of these
words with one  of its synonyms chosen at random.  In Random Insertion
(RI), we find a  random synonym of a random word  in the sentence that
is not a  stopword.  We insert that synonym into  a random position in
the sentence. We can do this n times. In Random Swap (RS), we randomly
choose two words in the sentence  and swaps their positions. We can do
this n times. In Random Deletion,  we randomly remove each word in the
sentence  with  probability  p.   On five  text  classification  tasks
(Stanford       Sentiment       Treebank,      Customer       reviews,
Subjectivity/Objectivity, TREC (Question type dataset) and PC (Pro-Con
dataset)),  they   show  that   EDA  improves  performance   for  both
convolutional   and  recurrent   neural  networks.   EDA  demonstrated
particularly  strong results  for  smaller datasets.   On an  average,
across five datasets,  training with EDA while using only  50\% of the
available training set  achieved the same accuracy  as normal training
with all available data.\\

Akbar et  al. \cite{AEDA} propose  AEDA (An Easier  Data Augmentation)
technique  to  help improve  the  performance  on text  classification
tasks. AEDA includes only random  insertion of punctuations marks into
the original text.  This is an  easier technique to implement for data
augmentation than EDA \cite{eda} method  with which they compare their
results. In addition,  it keeps the order of the  words while changing
their  positions  in the  sentence  leading  to a  better  generalized
performance. Furthermore, the deletion operation in EDA can cause loss
of  information which,  in turn,  misleads the  network, whereas  AEDA
preserves  all  the  input  information.  They  show  that  using  the
AEDA-augmented data for training, the models show superior performance
compared to  using the EDA-augmented  data in all five  datasets (SST2
Stanford Sentiment  Treebank \cite{sst2}, CR Customer  Reviews Dataset
\cite{hu2004},  SUBJ  Subjectivity/Objectivity  \cite{pang2004},  TREC
Question  Classification dataset  \cite{li2002}, PC  \cite{murthy2008}
Pros and Cons dataset). \\

Rashid  et al.  \cite{Mlm}  modify EDA  with  more sophisticated  text
editing  operations powered  by masked  language models  like BERT  or
RoBERTa  to  analyze  the  benefits   or  setbacks  of  creating  more
linguistically    meaningful    and    hopefully    higher    quality
augmentations.  Their  analysis  demonstrates   that  using  a  masked
language  model  for  word  insertion almost  always  achieves  better
results than the  initial method but it  comes at a cost  of more time
and  resources which  can  be comparatively  remedied  by deploying  a
lighter and smaller language model.\\

Zhang et al. \cite{enhanced} showed  the whole process of building the
WCC-EC  (Web  Common  Crawled  English-Chinese)  corpus  and  made  it
available for free  download.  In the process of  building the corpus,
they proposed a method to improved the alignment accuracy of bilingual
texts by combining paragraph  information.  Finally, they proposed a
simple  and  effective  method  for data  augmentation  by  retrieving
parallel partial  sentences from  parallel texts. This  approach could
improve  the quality  of the  translation model  without changing  the
translation  model architecture  and  was applicable  to all  language
pairs that support bilingual embedding.\\

Steven  et al.  \cite{Survey} present  a comprehensive  and structured
survey   of  data   augmentation  for   natural  language   processing
(NLP). They provide a good background  on data augmentation and how it
works,   and  discuss   major  methodologically   representative  data
augmentation  techniques for  NLP. They  touch upon  data augmentation
techniques for popular  NLP applications and tasks.  They also outline
current challenges and directions for future research.\\

Jiaao  et  al.  \cite{Chen}  provide  an  empirical survey  of  recent
progress  on data  augmentation for  NLP in  the limited  labeled data
setting, summarizing  the landscape of methods  (including token-level
augmentations, sentence-level augmentations, adversarial augmentations
and  hidden-space augmentations)  and carrying  out experiments  on 11
datasets  covering   topics/news  classification,   inference  tasks,
paraphrasing  tasks  and  single-sentence tasks.  They  trained  their
models on NVIDIA 2080ti and  NVIDIA V-100 GPUs. Supervised experiments
took 20 minutes, and semi-supervised experiments took two hours.\\

Tasnim  et  al.   \cite{AuGVIC}   have  presented  an  in-domain  data
augmentation framework  AUGVIC by  exploiting the bitext  vicinity for
low-resource   NMT.  Their   method  generates   vicinal  samples   by
diversifying sentences of the target language in the bitext in a novel
way. It  is simple yet  effective and can  be quite useful  when extra
in-domain monolingual data is limited.\\

Alexis et al. \cite{xlmr}  show that pretraining multilingual language
models leads to  significant performance gains for a wide
range of cross-lingual transfer  tasks. They train a Transformer-based
masked language  model on 100  languages using only  monolingual data,
using  more than  two terabytes  of filtered  CommonCrawl data.  Their
model,  dubbed  XLM-R,  significantly  outperforms  multilingual  BERT
(mBERT) on  a variety  of cross-lingual benchmarks,  including +14.6\%
average accuracy on  XNLI, +13\% average F1 score on  MLQA, and +2.4\%
F1  score on  NER. XLM-R  performs particularly  well on  low-resource
languages, improving  15.7\% in XNLI  accuracy for Swahili  and 11.4\%
for  Urdu over  previous XLM  models. They  claim XLM-RoBERTa  (XLM-R)
outperforms  mBERT  on  cross-lingual  classification by  up  to  23\%
accuracy on low-resource languages.  It outperforms the previous state
of the art by 5.1\% average  accuracy on XNLI, 2.42\% average F1-Score
on   Named  Entity   Recognition,  and   9.1\%  average   F1-Score  on
cross-lingual Question Answering.\\

Feng  et al.   \cite{feng} introduce  Language agnostic  BERT Sentence
Embedding (LaBSE), and compare  with LASER (Language Agnostic Sentence
Embedding  Representations)\cite{artetxe2019} and  m-USE (Multilingual
Unsupervised  and Supervised  Embeddings) \cite{yang2019}  approaches.
LaBSE  outperformed LASER  and m-USE  in many  scenarios.  LaBSE  is a
multilingual  sentence embedding  model  for more  than 109  languages
based on dual encoder transformer architecture of BERT [10,27]. The
model  achieves   state-of-the-art  performance  on   various  bi-text
retrieval/mining  tasks  compared  to the  previous  state-of-the-art,
while  also  providing increased  language  coverage.  The encoder  is
pre-trained with Masked Language  Model (MLM) and Translation Language
Model (TLM) trained on the  monolingual data and bilingual translation
pairs,  respectively.  Pre-training  uses TPUv3  with 512-cores  and a
batch size of 8192. The max sequence  length is set to 512 and 20\% of
tokens (or 80 tokens at most) per  sequence are masked for MLM and TLM
predictions.   For  the three  stages  of  progressive stacking,  they
respectively  train  for   400k,  800k,  and  1.8M   steps  using  all
monolingual and bilingual data.  Compared to LASER \cite{artetxe2019},
their models  perform significantly better on  low-resource languages,
boosting  the overall  accuracy on  112 languages  to 83.7\%  from the
65.5\%  achieved  by  the previous  state-of-art.  Surprisingly,  they
observe that their  models performs well on 30+  Tatoeba languages for
which  we have  no explicit  monolingual or  bilingual training  data.
LaBSE not only  systematically outperforms prior work  but also covers
all languages within a single model.\\

Tharindu et al.   \cite{Tharindu} propose a simple  QE framework based
on cross-lingual transformers and we  use it to implement and evaluate
two   different  neural   architectures   to  perform   sentence-level
architectures. The first architecture proposed, MonoTransQuest, uses a
single  XLM-R  transformer  model.   The  input of  this  model  is  a
concatenation of the original  sentence and its translation, separated
by the [SEP] token. They  experiment with three pooling strategies for
the output  of the transformer  model: using  the output of  the [CLS]
token (CLS-strategy); computing the mean  of all output vectors of the
input  words (MEAN-strategy);  and  computing a  max-over-time of  the
output vectors  of the input  words (MAX-strategy). The output  of the
pooling strategy is used as the input of a softmax layer that predicts
the  quality score  of the  translation. They  used mean-squared-error
loss as  the objective function.  Early  experiments demonstrated that
the CLS-strategy leads to better results than the other two strategies
for  this architecture.   Therefore, they  used the  embedding of  the
[CLS] token  as the  input of  a softmax  layer.  The  Second approach
proposed, SiameseTransQuest,  based on the Siamese  architecture, gave
promising  results in  monolingual semantic  textual similarity  tasks
\cite{Reimers} .   In this case, they  feed the original text  and the
translation into two separate XLM-R transformer models. Similar to the
previous architecture they used the  same three pooling strategies for
the outputs of the transformer models. They then calculated the cosine
similarity between the two outputs of the pooling strategy.  They used
mean-squared-error  loss  as  the   objective  function.   In  initial
experiments they carried out with this architecture, the MEAN-strategy
showed better results than the other two strategies.  For this reason,
they  used the  MEAN-strategy  for their  further experiments.  Cosine
similarity is calculated between the the mean of all output vectors of
the  input  words produced  by  each  transformer.   For some  of  the
experiments, they used an Nvidia Tesla K80 GPU, whilst for others they
used an Nvidia Tesla T4 GPU.   On an Nvidia Tesla K80 GPU, MTransQuest
takes 4,480s on average to train on 7,000 instances, while STransQuest
takes only 3,900s on average for the same experiment. On the same GPU,
MTransQuest  takes  35s  on  average to  perform  inference  on  1,000
instances which takes STransQuest only 16s to do so.\\

Xuan-Phi et al. \cite{Nguyen} introduce Data Diversification: a simple
but  effective  strategy to  boost  neural  machine translation  (NMT)
performance. It diversifies the training data by using the predictions
of multiple forward and backward models and then merging them with the
original dataset on which the final  NMT model is trained. It does not
require extra monolingual data like  back-translation, nor does it add
more computations and parameters like  ensemble of models. Their method
achieves state-of-the-art  BLEU Scores of  30.7 and 43.7 in  the WMT14
English-German   and  English-French   translation  tasks.    It  also
substantially improves  on 8  other translation  tasks: 4  IWSLT tasks
(English-German  and English-French)  and  4 low-resource  translation
tasks (English-Nepali and English-Sinhala).\\

Nandan  Thakur  et al.   \cite{Reimers}  present  a data  augmentation
method, which they  call Augmented SBERT (AugSBERT), that  uses a BERT
cross-encoder  to improve  the performance  for the  SBERT bi-encoder.
They conduct experiments using  PyTorch Huggingface's Transformers and
the  sentence-transformers framework  \cite{Nils}.  The latter  showed
that  BERT outperforms  other transformer-like  networks when  used as
bi-encoder.  For English datasets,  they use bert-base-uncased and for
the  Spanish  dataset  they use  bert-base-multilingual-cased.   Every
AugSBERT model  exhibits computational  speeds identical to  the SBERT
model \cite{Nils}.\\

Jain  et  al.   \cite{jain}   implement  various  English-Marathi  and
Marathi-English  baseline NMT  systems and  use the  given monolingual
Marathi  data to  implement  the back-translation  technique for  data
augmentation.  They  also discuss the  details of the  various Machine
Translation   (MT)   system  that   they   have   submitted  for   the
English-Marathi  LoResMT task.   They have  submitted three  different
Neural Machine Translation (NMT)  systems; a Baseline English-Marathis
system,  a  Baseline  Marathi-English System  and  an  English-Marathi
System that is  based on the back-translation  technique. They explore
the  performance of  these  NMT systems  between  English and  Marathi
languages,  which   forms  a  low   resource  language  pair   due  to
unavailability  of sufficient  parallel data.   They also  explore the
performance of the back-translation technique when the back-translated
data is  obtained from NMT  Systems that are  obtained on a  very less
amount   of  data.   They  train   a  NMT   System  using   the  given
English-Marathi  parallel corpus  for 1600  epochs and  they save  the
model  for   every  200  epochs   starting  from  200,  to   test  the
performance.  They   have  used  the   transformer-based  architecture
consisted of  6 encoder  layers and  6 decoder  layers. The  number of
encoder and decoder attention heads used were 4 each. They use encoder
and decoder embedding dimension of 512 each.  For Training the system,
the  optimizer used  was Adam  optimizer with  betas (0.9,  0.98). The
inverse  square root  learning rate  scheduler was  used with  initial
learning  rate   of  5e-4  and  4,000   warm-up  updates.The  baseline
English-Marathi system  produced a BLEU  score of 11 and  the baseline
Marathi-English  System produced  a  BLEU score  of  17.2.  They  also
observe  that the  English-Marathi system  gives the  best BLEU  score
after  400 epochs  and after  that the  scores decrease  and fluctuate
between a small range. The  Marathi-English model gives the best score
after 600 epochs and after that the scores starts decreasing.\\

Banerjee et  al. \cite{banerjee2021}  explore different  techniques of
overcoming   the  challenges   of  low-resource   in  neural   Machine
Translation (NMT), especially focusing  on the case of English-Marathi
NMT.  They try  to  mitigate the  low-resource  problem by  augmenting
parallel corpora  or by  using transfer  learning. Techniques  such as
Phrase Table Injection (PTI),  back-translation and mixing of language
corpora are used for enhancing the parallel data, whereas pivoting and
multilingual embeddings  are used to leverage  transfer learning.  For
pivoting,  Hindi comes  in as  assisting language  for English-Marathi
translation. They  have shown that  the pivot based  transfer learning
approach can significantly improve  the quality of the English-Marathi
translations  over  the  baseline  by  using  Hindi  as  an  assisting
language. They also observe that the phrases from the SMT training can
help the NMT  model perform better. The one (English)  to many (Hindi,
Marathi)  multilingual model  is able  to improve  the English-Marathi
translations by leveraging the English-Hindi parallel corpus.\\

Goyal  et  al. \cite{goyal2020}  propose  a  technique called  Unified
Transliteration   and  Subword   Segmentation  to   leverage  language
similarity  while  exploiting  parallel  data  from  related  language
pairs. They also propose a Multilingual Transfer Learning technique to
leverage  parallel  data from  multiple  related  languages to  assist
translation for low resource language pair of interest.\\

Khatri  et al.   \cite{khatri2021} describe  their submission  for the
shared task on Unsupervised MT and  Very Low Resource Supervised MT at
WMT 2021.   They submitted systems  for two language pairs:  German to
Upper Sorbian  (de to hsb)  and Upper Sorbian  to German (hsb  to de),
German to Lower  Sorbian (de to dsb) and Lower  Sorbian to German (dsb
to de).  For de to  hsb, hsb to de,  they pretrain their  system using
MASS (Masked Sequence to Sequence)  objective and then fine-tune using
iterative back-translation. For  de to dsb and dsb to  de, no parallel
data is provided in this task, they use  final de to hsb and hsb to de
model as initialization of the de to dsb and dsb to de model and train
it further using iterative back-translation, using the same vocabulary
as used in the de to dsb and dsb to model.\\

Himmet  et  al. \cite{Iterative}  propose  a  novel text  augmentation
method that  leverages the Fill-Mask feature  of the transformer-based
BERT  model. Their  method  involves iteratively  masking  words in  a
sentence and replacing them with language model predictions. They have
tested their proposed  method on various NLP tasks and  found it to be
effective in many cases. Experimental results show that their proposed
method  significantly   improves  performance,  especially   on  topic
classification datasets.

\section{Methodology}

We start  with a  seed corpus.  This will be  a parallel  corpus. This
corpus will include two plain  text files, one containing sentences in
L1 and other containing the corresponding translations in L2. For each
sentence pair s1-s2 in this corpus, we do the following:\\

Mask a  word and  replace it  with other words  predicted by  a Masked
Language Model  such as  XLM-R-Base or  XLM-R-Large.  For  each masked
word,  several  replacement  words  will be  predicted,  resulting  in
several new sentences. The whole idea  of replacing only one word at a
time is  that syntax and  semantics are  least disturbed when  we make
only small changes at a time.\\

This way,  we generate  new sentences  starting from  a given  pair of
sentences  s1-s2, where  s1 is  in L1  and s2  is in  L2. Some  of the
sentences so generated in L2  may be translational equivalents of some
of  the sentences  generated  in L1.   If  we had  a  good word  level
alignment technique,  we could  have limited our  focus to  only those
sentences  generated by  masking aligned  words. In  the absence  of a
reliable alignment technique, we  compare each sentence generated from
s1 with  each sentence  generated from s2,  and choose  those sentence
pairs which are  likely to be translations of each  other.  This we do
by  comparing  Sentence  Embedding scores.   Sentence  Embeddings  are
supposed to  capture the semantics  of sentences and if  two sentences
are  translations   of  one  another,  they   should  be  semantically
similar. We can use LaBSE, LASER etc. for this purpose.\\

Let us  illustrate the whole  process with  a single sentence  pair in
English and Hindi languages:\\

\begin{figure}
  \includegraphics[width=0.99\columnwidth]{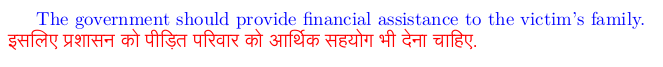}
  \caption{Original English-Hindi Sentences}
  \label{fig:example}
\end{figure}
By masking different words and  predicting alternatives for the masked
words, let us say  we generate  50 new  sentences in  English  and 60  new
sentences in Hindi. This gives us a total of 3000 sentence pairs. Some
of these generated sentences are shown below:\\

\begin{figure}
  \includegraphics[width=0.99\columnwidth]{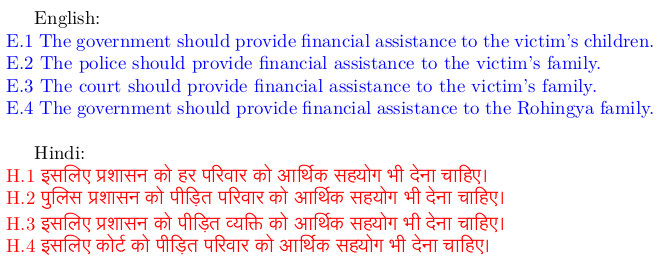}
  \caption{Generated English and Hindi Sentences}
  \label{fig:example}
\end{figure}

Thus we get 16 sentence pairs  from these generated sentences. Some of
these pairs will  be good translations of each other  while others may
not be.  For example, E.3 and  H.4 are good translations of each other
while E.1  and H.1 are  not.  How do we  then select the  right pairs?
E.3 was  obtained by  masking the word  {\color{blue}'government'} and
replacing it  with the word  {\color{blue}'court'}.  In H.4,  the word
{\color{red}'prashaasan'}  in  Romanized  form  has  been  masked  and
replaced with the Romanized  word {\color{red}'court'}.  The Romanized
Hindi word  {\color{red}'prashaasan'} is a translation  of the English
word    {\color{blue}'government'}    and     the    Romanized    word
{\color{red}'court'}    is   the    same   as    the   English    word
{\color{blue}'court'}.  Other words in the original sentences have not
been touched. Therefore, E.3 and H.4 can be taken as good translations
of each  other.  H1  means every family  should be  provided financial
assistance by the  government, which is not the meaning  of E.1. Given
this basic  idea, we can think  of several methods of  identifying the
correct pairs  of sentences  for inclusion  in the  augmented parallel
corpus.  If  we had  a good bilingual  dictionary, which  includes not
only root words  but also all the morphologically  inflected words, we
could  simply  use  such  a bilingual  dictionary  to  pick  generated
sentence  pairs. In  the absence  of such  a bilingual  dictionary, we
could  build a  Word Co-Occurrence  Matrix \cite{kumari2023}  from the
seed  or  other parallel  corpora  available  for  the given  pair  of
languages.  When the co-occurrence frequency is high, the co-occurring
words tend to be good translations of  each other and this idea can be
used  for  the present  purpose.   Thirdly,  if  we  had a  good  word
alignment algorithm, we  could use that, but in practice,  none of the
available word alignment techniques are  reliable enough.  As a fourth
alternative, we  could use  sentence embeddings  and check  the cosine
similarity score for each of the  generated sentence pairs and pick up
sentences with a  good similarity score.  We give  below more examples
of good and bad sentence pairs with corresponding cosine similarity of
LaBSE scores:\\

\begin{figure}
  \includegraphics[width=0.99\columnwidth]{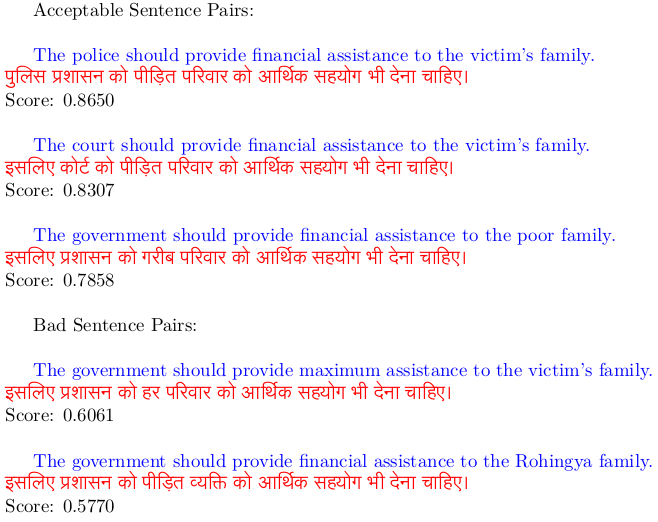}
  \caption{Acceptable and bad Sentence Pairs}
  \label{fig:example}
\end{figure}

It can seen  that sentence pairs which are translations  of each other
tend to get  higher similarity score compared to  sentence pairs which
are not. This is the core idea of this paper.\\

By repeating this procedure on all  the sentence pairs in the parallel
corpus, we  can generate a  much larger parallel corpus,  hopefully of
reasonably good quality.\\

Once the entire seed corpus is processed and augmented parallel corpus
is generated,  we can repeat the  whole process to build  still larger
parallel corpora with greater diversity.  For example, we can generate
{\color{blue}'The hospital  should provide financial assistance  to the
  victim’s  family'}   from  {\color{blue}'The  court   should  provide
  financial assistance to the  victim’s family'}, and {\color{blue}'The
  hospital should provide medical  assistance to the victim’s family'}
from that  and then  {\color{blue}'The hospital should  provide medical
  assistance to the poor family'} and so on.\\

This methodology will work for all languages for which Masked Language
Models are available.\\

We can exclude stop words  (including function words) so that sentence
structure is not drastically affected by the process.\\

The augmented  corpus so generated  can be cross-checked using  MT Quality
Estimation methods.

\section{Experiments and Results}

We use XLM-RoBERTa for masking and predicting alternative words to fit
the masked word. It is trained on 2.5TB of filtered CommonCrawl data in
100 languages. The details of the various XLM-RoBERTa models are given
below.\\

XLM-R  models are  the largest  models.  The parameters  of the  XLM-R
models are 250K  vocabulary, number of layers is 24,  number of hidden
states of the  model is 1024, the dimension of  the feed-forward layer
Hff is  4096, number of  Attention heads A is  16 and total  number of
parameters is 550 Million.\\

Configuration  details of  multilingual XLM-R-base  models are:  250k
vocabulary ,  number of layers is  12, number of hidden  states of the
model Hm is 768, the dimension  of the feed-forward layer Hff is 3072,
number of  Attention heads  A is  12 and total  of parameters  are 270
Million.\\

Details of all  the variants are shown in Table  1. In our experiments
we use XLM-R-base.\\

The LaBSE  \cite{feng} model is  a Dual encoder model  with BERT-based
encoding  modules   to  obtain  sentence  embeddings.    The  sentence
embeddings   are  extracted   as   the  l2   normalized  [CLS]   token
representations from the last transformer block.\\

We first extract sentence embeddings  for the sentences generated from
the L1 sentence  and sentence embeddings for  sentences generated from
the L2 sentence.  We then  compute cosine similarity score between the
source  and the  corresponding target  sentence embeddings.  We select
sentence pairs with cosine similarity score above a threshold.\\

We then use a MT Quality Estimation tool TransQuest \cite{Tharindu} to
cross check the augmented corpus so generated.\\

\textbf{TransQuest  Scoring:} We  feed “s[SEP]t"  as an  input to  the
MonoTransQuest \cite{Tharindu} architecture which  uses a single XLM-R
model. Here s and  t stand for L1 and L2 sentences.  The output of the
[CLS] token is used as the input  of a softmax layer that predicts the
quality score of the s-t sentence pair

\indent\begin{table}
\centering{}%
\begin{tabular}{|l|l|l|l|l|}
\hline

\textbf{Models}       & \textbf{\# Layers}     	  & \textbf{Model Dimension} & \textbf{Vocabulary} &\textbf{Parameters}  \\
\hline
\textbf {xlmr.base}  & 12	  & 3072  & 250k    & 270M   \\
\textbf {xlmr.large} & 24	  & 4096  &  250k &  550M\\	
\textbf{xlmr.xl}	   & 36   & 2560  & 250k &   3.5B\\	
\textbf{xlmr.xxl}   & 48   & 4096  & 250k  &  10.7B\\	     
\hline
\end{tabular}
\noindent\caption{Configuration on Masked Language Model (XLM-RoBERTa) \cite{xlmr,xlmr-xl}}
\end{table}

\textbf{Experiment 1}

Given a single pair  of sentences in English and Hindi,  let us say we
mask  5 words  in the  English sentence  and generate  50 new  English
sentences by setting the hyper-parameter topk as 10. Similarly we mask
6 words  in Hindi, say,  and generate 60  new sentences.  Thus  we get
3000  sentence pairs.  Of these  3000 sentences,  200 sentences  get a
LaBSE  score of  greater  than  equal to  0.80.   This  way, we  could
generate  200 new  sentence pairs  of reasonably  good quality  from a
single sentence pair in the seed corpus.  By running this process over
the whole seed corpus, we can  generate a hundred times bigger corpus.
We could  repeat the process on  the augmented corpus so  generated to
produce even larger corpora.\\

Manual observation  of samples so  generated shows that  the generated
augmented  corpus  is of  reasonably  good  quality. Further,  we  run
TransQuest on the 200 generated  sentence pairs mentioned above and we
get a score of above 0.8 in all cases, thereby  further validating the
efficacy of the proposed method.

\section{Conclusions}

Here  we have  proposed a  simple and  effective method  of generating
augmented parallel corpora from a given seed corpus. Compared to other
techniques  such  as  manipulating  punctuation  marks,  splitting  of
sentences etc. to generate augmented corpus, our approach can generate
syntactically and semantically better  quality sentences with enhanced
diversity.   We believe  this method  can greatly  alleviate the  data
scarcity problem for  all language pairs for which  a reasonable sized
seed corpus is available.

\bibliography{par-corp-aug}

\begin{thebibliography}{10}

\bibitem{artetxe2019}
Artetxe, Mikel Schwenk, and Holger.
\newblock Massively multilingual sentence embeddings for zero-shot
  cross-lingual transfer and beyond.
\newblock {\em Transactions of the association for computational linguistics},
  7:597--610, 2019.

\bibitem{banerjee2021}
Aakash Banerjee, Aditya Jain, Shivam Mhaskar, Sourabh Deoghare, Aman Sehgal,
  and Pushpak Bhattarchrayya.
\newblock Neural machine translation in low-resource setting: a case study in
  english-marathi pair.
\newblock In {\em Proceedings of Machine Translation Summit XVIII: Research
  Track}, pages 35--47, 2021.

\bibitem{Brown}
Brown and Tom B.
\newblock Language models are few-shot learners.
\newblock In {\em Advence in neural Information Processing System}, pages
  1877--1901, 2020.

\bibitem{Chen}
Jiaao Chen, Derek Tam, Colin Raffel, Mohit Bansal, and Diyi Yang.
\newblock An empirical survey of data augmentation for limited data learning in
  nlp.
\newblock In {\em Transactions of the Association for Computational
  Linguistics}, pages 191--211, 2023.

\bibitem{xlmr}
Alexis Conneau, Kartikay Khandelwal, Naman Goyal, Vishrav Chaudhary, Guillaume
  Wenzek, Francisco Guzman, Edouard Grave, Myle Ott, Luke Zettlemoyer, and
  Veselin Stoyanov.
\newblock Unsupervised cross-lingual representation learning at scale.
\newblock In {\em Proceedings of the 58th Annual Meeting of the Association for
  Compuatational Linguistics}, pages 8440--8451, 2020.

\bibitem{kenton}
Jacob Devlin, Ming-Wei Chang, Lee Kenton, and Kristina Toutanova.
\newblock Bert: Pre-training of deep bidirectional transformers for language
  understanding.
\newblock In {\em Proceedings of naacL-HLT}, volume~1, page~2, 2019.

\bibitem{feng}
Fangxiaoyu Feng, Yinfei Yang, Daniel Cer, Naveen Arivazhagan, and Wei Wang.
\newblock Language-agnostic bert sentence embeddings.
\newblock In {\em Proceeding of the 60th Annual Meeting of the Association for
  Computational Linguistics}, pages 878--891, 2022.

\bibitem{Survey}
Steven~Y. Feng, Varun Gangal, Jason Wei, Sarath Chandar, Soroush Vosoughi,
  Teruko Mitamura, and Eduard Hovy.
\newblock A survey of data augmentation aproaches for nlp.
\newblock In {\em Findings of the Association for Computational Linguistics:
  ACL-IJCNLP 2021}, pages 968--988, 2021.

\bibitem{firata}
Orhan Firat, Kyunghyun Cho, and Yoshua Bengio.
\newblock Multi-way, multilingual neural machine translation with a shared
  attention mechanism.
\newblock In {\em In Proceedings of the 2016 Conference of the North American
  Chapter of the Association for Computational Linguitics: Human Language
  Technologies}, pages 866--875, 2016a.

\bibitem{firatb}
Orhan Firat, Baskaran Sankaran, Yaser Al-onaizan, Fatos T.~Yarman Vurala, , and
  Kyunghyun Cho.
\newblock Zero-resource translation with multi-lingual neural machine
  translation.
\newblock In {\em In proceedings of the 2016 Conference on Empirical Methods in
  Natural Language Processing}, pages 268--277, 2016b.

\bibitem{murthy2008}
Murthy Ganapathibhotla and Bing Liu.
\newblock Mining opinions in comparative sentences.
\newblock In {\em Proceedings of the 22nd international conference on
  computational linguistics (Coling 2008)}, pages 241--248, 2008.

\bibitem{xlmr-xl}
Naman Goyal, Jingfei Du, Myle Ott, Giri Anantharaman, and Alexis Conneau.
\newblock Large-scale transformers for multilingual masked language modeling.
\newblock In {\em Proceedings of the 6th Workshop on Representation Learning
  for NLP (RepL4NLP-2021)}, pages 29--33, 2021.

\bibitem{goyal2020}
Vikrant Goyal, Sourav Kumar, and Dipti~Misra Sharma.
\newblock Efficient neural machine translation for low-resource languages via
  exploiting related languages.
\newblock In {\em Proceedings of the 58th annual meeting of the association for
  computational linguistics: student research workshop}, pages 162--168, 2020.

\bibitem{hayashi}
Tomoki Hayashi, Shinji Watanabe, Yu~Zhang, Tomoki Toda, Takaaki Hori, Ramon
  Astudillo, and Kazuya Takeda.
\newblock Back-translation-style data augmentation for end-to-end asr.
\newblock In {\em 2018 IEEE Spoken Language Technology Workshop (SLT)}, pages
  426--433. IEEE, 2018.

\bibitem{He}
Junxian He, Jiatao Gu, Jiajun Shen, and Marc’Aurelio Ranzato.
\newblock Revisting self-training for neural sequence generation.
\newblock In {\em ICLR}, 2020.

\bibitem{hu2004}
Minqing Hu and Bing Liu.
\newblock Mining and summarizing customer reviews.
\newblock In {\em Proceedings of the tenth ACM SIGKDD international conference
  on Knowledge discovery and data mining}, pages 168--177, 2004.

\bibitem{jain}
Aditya Jain, Shivam Mhaskar, and Pushpak Bhattacharyya.
\newblock Evaluating the performance of back-translation for low resource
  english-marathi language pair: Cfilt-iitbombay@ loresmt 2021.
\newblock In {\em Proceedings of the 4th workshop on technologies for MT of low
  resource languages (LoResMT2021)}, pages 158--162, 2021.

\bibitem{AEDA}
Akbar Karimi, Leonardo Rossi, and Andrea Prati.
\newblock Aeda: An easier data augmentation technique for text classification.
\newblock In {\em Findings of the Association for Computational Linguistics:
  EMNLP 2021}, pages 2748--2754, 2021.

\bibitem{Iterative}
Himmet~Toprak Kesgin and Mehmet~Fatih Amasyali.
\newblock Iterative mask filling: An effective text augmentation method using
  masked language modeling.
\newblock In {\em ICAETA, 2023}, pages 450--463, 2024.

\bibitem{khatri2021}
Jyotsana Khatri, Rudra~Murthy V, and Pushpak Bhattacharyya.
\newblock Language model pretraining and transfer learning for very low
  resource languages.
\newblock In {\em Proceedings of the Sixth Conference on Machine Translation},
  pages 995--998, 2021.

\bibitem{Koehn}
Philipp Koehn and Rebecca Knowles.
\newblock Six challenges for neural machine translation.
\newblock In {\em Proceedings of the First Workshop on Neural Machine
  Translation}, pages 28--39, 2017.

\bibitem{kumari2023}
Vibhuti Kumari and Kavi~Narayana Murthy.
\newblock Quality estimation of machine translated texts based on direct
  evidence from training data.
\newblock {\em arXiv preprint arXiv:2306.15399}, 2023.

\bibitem{li2002}
Xin Li and Dan Roth.
\newblock Learning question classifiers.
\newblock In {\em COLING 2002: The 19th International Conference on
  Computational Linguistics}, 2002.

\bibitem{marivate}
Vukosi Marivate and Tshephisho Sefara.
\newblock Improving short text classification through global augmentation
  methods.
\newblock In {\em Machine Learning and Knowledge Extraction: 4th IFIP TC 5, TC
  12, WG 8.4, WG 8.9, WG 12.9 International Cross-Domain Conference, CD-MAKE
  2020, Dublin, Ireland, August 25--28, 2020, Proceedings 4}, pages 385--399.
  Springer, 2020.

\bibitem{mikolov}
Tomas Mikolov, Kai Chen, Greg Corrado, and Jeffrey Dean.
\newblock Efficient estimation of word representations in vector space.
\newblock {\em arXiv preprint arXiv:1301.3781}, 2013.

\bibitem{AuGVIC}
Tasnim Mohiuddin, M~Saiful Bari, and Shafiq Joty.
\newblock Augvic: Exploiting bitext vicinity for low-resource nmt.
\newblock In {\em Findings of the Association for Computational Linguistics:
  ACL-IJCNLP}, pages 3034--3045, 2021.

\bibitem{Nguyen}
Xuan-Phi Nguyen, Shafiq~R. Joty, Kui Wu, and Ai~Ti Aw.
\newblock Data diversification: A simple strategy for neural machine
  translation.
\newblock In {\em In Advances in Neural Information Processing Systems 33:
  Annual Conference on Neural Information Processing Systems}, pages 6--12,
  2020.

\bibitem{pang2004}
Bo~Pang and Lillian Lee.
\newblock A sentimental education: Sentiment analysis using subjectivity
  summarization based on minimum cuts.
\newblock {\em arXiv preprint cs/0409058}, 2004.

\bibitem{Tharindu}
Tharindu Ranasinghe, Constantin Orasan, and Ruslan Mitkov.
\newblock Transquest: Translation quality estimation with cross-lingual
  transformers.
\newblock In {\em Proceedings of the 28th International Conference on
  Computational Linguistics}, pages 5070--5081, 2020.

\bibitem{Mlm}
Mohammad~Amin Rashid and hossein Ammirkhani.
\newblock The effect of using masked language models in random textual data
  augmentation.
\newblock In {\em 26th International Computer Conference, Computer Society of
  Iran, Tehran, Iran}, 2021.

\bibitem{Nils}
Nils Reimers and Iryna Gurevych.
\newblock Sentence-bert: Sentence embeddings using siamese bert-networks.
\newblock In {\em Proceedings of the 2019 Conference on Empirical Methods in
  Natural Language Processing and the 9th International Joint Conference on
  Natural Language Processing}, pages 3982--3992, 2019.

\bibitem{Georgios}
Georgios Rizos, Konstantin Hemker, and Bjorn Schuller.
\newblock Augment to prevent: short-text data augmentation in deep learning for
  hate-speech classification.
\newblock In {\em Proceedings of the 28th ACM International Conference on
  Information and Knowledge Management}, pages 991--1000, 2019.

\bibitem{Sennrich}
Rico Sennrich, Barry Haddow, and Alexandra Birch.
\newblock Improving neural machine translation models with monolingual data.
\newblock In {\em Proceedings of the 54th Annual Meeting of the Association for
  Computational Linguistics}, pages 86--96, 2016.

\bibitem{Connor}
Connor Shorten, Taghi~M. Khoshgoftaar, and Borko Furht.
\newblock Text data augmentation for deep learning.
\newblock {\em Journal of Big Data}, 2021.

\bibitem{sst2}
Richard Socher, John Bauer, Christopher~D Manning, and Andrew~Y Ng.
\newblock Parsing with compositional vector grammars.
\newblock In {\em In Proceedings of the 51st Annual Meeting of the Association
  for Computational Linguistics}, pages 455--465, 2013.

\bibitem{Reimers}
Nandan Thakur, Nils Reimers, Johannes Daxenberger, and Iryna Gurevych.
\newblock Augmented sbert: Data augmentation method for improving bi-encoders
  for pairwise sentence scoring tasks.
\newblock In {\em Proceedings of the 2021 Conference of the North American
  Chapter of the Association for Computational Linguistics: Human Language
  Technologies}, pages 296--310, 2021.

\bibitem{vaswani2017}
Ashish Vaswani, Noam Shazeer, Niki Parmar, Jakob Uszkoreit, Llion Jones,
  Aidan~N. Gomez, Lukasz Kaiser, and Illia Polosukhin.
\newblock Attention is all you need.
\newblock {\em Advances in Neural Information Processing Systems}, 2017.

\bibitem{eda}
Jason Wei and Kai Zou.
\newblock Eda: Easy data augmentation techniques for boosting performance on
  text classification tasks.
\newblock In {\em Proceedings of the 2019 Conference on Empirical Methods in
  Natural Language Processing and the 9th International Joint Conference on
  Natural Language Processing}, pages 6382--6388, 2019.

\bibitem{yang2019}
Yinfei Yang, Daniel Cer, Amin Ahmad, Mandy Guo, Jax Law, Noah Constant,
  Gustavo~Hernandez Abrego, Steve Yuan, Chris Tar, Yun-Hsuan Sung, Brian
  Strope, and Ray Kurzweil.
\newblock Multilingual universal sentence encoder for semantic retrieval.
\newblock {\em arXiv preprint arXiv:1907.04307}, 2019.

\bibitem{enhanced}
Zhang, Jinyi Guo, Cong Mao, Jiannan Guo, Chong Matsumoto, and Tadahiro.
\newblock An enhanced method for neural machine translation via data
  augmentation based on the self-constructed english-chinese corpus, wcc-ec.
\newblock {\em IEEE Access}, 2023.

\end{thebibliography}

\end{document}